\documentclass[]{spie}  %>>> use for US letter paper
\usepackage{tabularray}
\usepackage{hhline}
\usepackage{colortbl}
\usepackage{amsmath,amsfonts,amssymb}
\usepackage{graphicx}
\usepackage{fixltx2e}
\usepackage{soul}
\usepackage{booktabs}
\usepackage{makecell}
\usepackage{float}
\usepackage{wrapfig}
\usepackage{multirow}
\usepackage[numbers, compress]{natbib}
\usepackage[table]{xcolor}
\usepackage[T1]{fontenc}
\usepackage{ulem}            % provides \uline - potentially better for simple underlining

\usepackage[colorlinks=true, allcolors=blue]{hyperref}

\title{SparseC-AFM: a deep learning method for fast and accurate characterization of MoS\textsubscript{2} with C-AFM }

\author[a]{Levi Harris}
\author[b]{Md Jayed Hossain}
\author[a]{Mufan Qiu}
\author[a]{Ruichen Zhang}
\author[b]{Pingchuan Ma}
\author[a]{Tianlong Chen}
\author[b]{Jiaqi Gu}
\author[c]{Seth Ariel Tongay}
\author[b]{Umberto Celano}

\affil[a]{Department of Computer Science, University of North Carolina, Chapel Hill}
\affil[b]{School of Electrical, Computer and Energy Engineering, Arizona State University}
\affil[c]{School for Matter, Transport and Energy Engineering, Arizona State University}

\authorinfo{Further author information: (Send correspondence to Levi Harris)\\Levi Harris: E-mail: levlevi@cs.unc.edu}

\begin{document} 
\maketitle

\begin{abstract}

The increasing use of two-dimensional (2D) materials in nanoelectronics demands robust metrology techniques for electrical characterization, especially for large-scale production. While atomic force microscopy (AFM) techniques like conductive AFM (C-AFM) offer high accuracy, they suffer from slow data acquisition speeds due to the raster scanning process. To address this, we introduce \textbf{SparseC-AFM}, a deep learning model that rapidly and accurately reconstructs conductivity maps of 2D materials like MoS$_2$ from sparse C-AFM scans. Our approach is robust across various scanning modes, substrates, and experimental conditions. Here we report on the comparison between (a.) classic flow implementation, where a high pixel density C-AFM image (i.e., 15min to collect) is manually parsed for the extraction of relevant materials parameters, and (b.) our SparseC-AFM method, where the same operation is performed with a dataset requiring substantially less time for acquisition (i.e., less than 5min). SparseC-AFM enables efficient extraction of critical material parameters in MoS$_2$, including film coverage, defect density, and identification of crystalline island boundaries, edges, and cracks. We achieved over \textbf{11$\times$ reduction in acquisition time} compared to the manual extraction of the same information from a full-resolution C-AFM image. Moreover, we demonstrate that our model-predicted samples have remarkably \textbf{similar electrical properties} to full-resolution data gathered using classic-flow scanning. This work represents a significant step towards translating AI-assisted 2D material characterization from laboratory research to industrial fabrication. All code and model weights are publicly available: \href{https://github.com/UNITES-Lab/sparse-cafm}{\texttt{https://github.com/UNITES-Lab/sparse-cafm}}

\end{abstract}
\section{Introduction}

Transition metal dichalcogenides (TMDs) are layered materials composed of MX$_2$ trilayers, where M is often Mo or W, and X is S, Se, or Te, characterized by strong covalent bonding between different atoms in-plane and weak van der Waals bonds between different layers. The attractive electronic and optoelectronic properties of TMDs have made them a frontrunner in research to develop ultra-thin body channel field effect transistors (FET) architectures. Substantial progress in dedicated TMD growth techniques has enabled the rapid fabrication of multiple samples that require characterization.

\subsection{C-AFM Material Characterization}

For example, material uniformity and defect control over large areas remain far below the standards of semiconductor manufacturers. Moreover, achieving wafer-scale probing of intrinsic material properties, extrinsic (process-induced) defects, and inhomogeneities remains a critical challenge, impeding the full potential of these devices \cite{obrien2023process}. Using high-resolution techniques for structural and compositional analysis, existing literature has reported on the relationships between electronic properties of TMDs and various process steps required for TMD synthesis and device fabrication \cite{sheng2023deep, martis2023imaging}. Examples include the effects of individual defects on optoelectronic properties under strain conditions and the correlation between point defects, grain boundaries (GBs), and wrinkles on electronic transport properties \cite{rosenberger2020twist, yang2014lattice, rosenberger2018electrical}. However, such methods are relatively time-consuming, and new methods that accelerate the ``lab-to-fab'' transition would offer enormous benefits for the metrology community.

Conductive atomic force microscopy (C-AFM) is a powerful characterization technique that operates at nanometric resolution, enabling morphological analysis and measurement of individual defects and creating a complete picture of deposited material quality. This approach involves raster scanning of a target area (Figure~\ref{fig:classic-flow}a-c), where the resulting image is the geometrical convolution between the tip-apex and surface features. Thus, there is a clear relationship between image quality and line density when using C-AFM. To clarify, consider imaging 10 nm features with a probe with a tip radius of 10 nm. Here, it is critical to set a line density of 5 nm/pixel or below to generate the correct image. Therefore, acquiring a dataset with a sufficient resolution for analysis of material properties such as defect density, coverage, or localization of grain boundaries, often requires \textit{10-20 minutes per image}. In conventional systems, attempts to reduce scanning time yield unacceptable image quality (Figure~\ref{fig:classic-flow}d).

\begin{figure}[tbp]
    \centering
    \includegraphics[width=1.0 \textwidth]{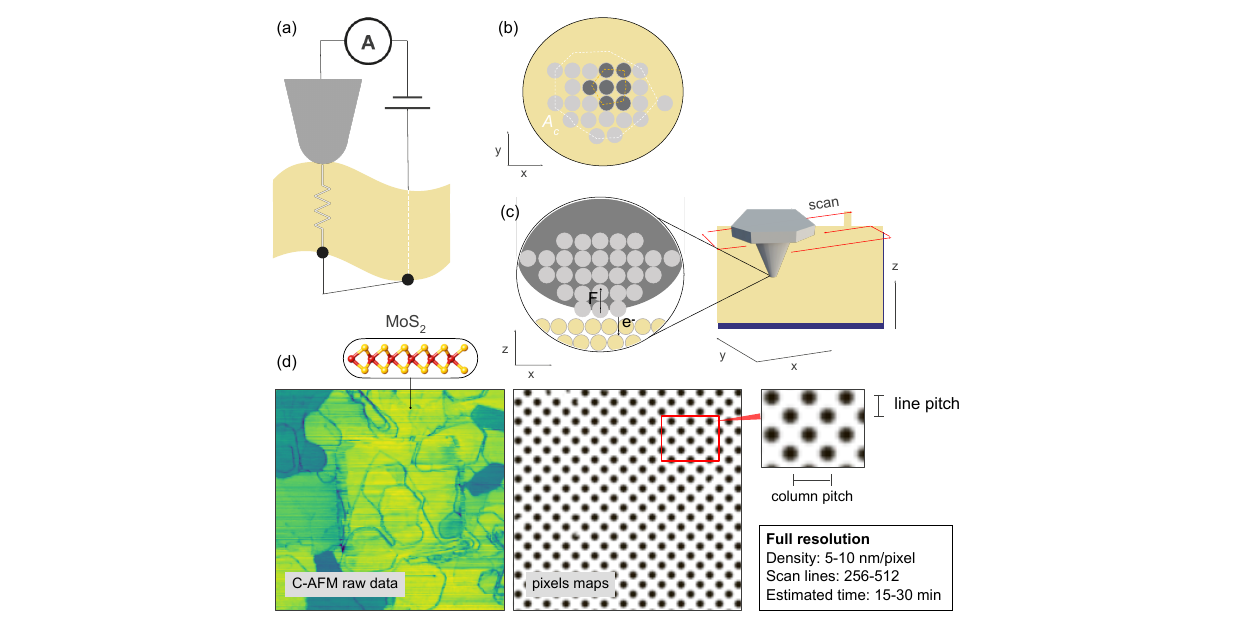}
    \caption{
        \textbf{Classic flow implementation} for full-resolution mapping of MoS$_2$ with C-AFM.
        \textbf{a)} Conductive AFM tip sweeps an MoS$_2$ surface as an amplifier (\textbf{A}) records current response under an applied bias.
        \textbf{b}) Path traced by a conductive tip; each dot represents a sampling point on a dense grid.
        \textbf{c}) Cross-section of the junction between conductive tip and material surface. $\sim$10 nm-radius tip (grey circles) makes contact with MoS$_2$ lattice (yellow circles) with normal loading force $F$ and current path $e^{-}$.
        \textbf{d}) Full-resolution dataset produced by classic-flow implementation. 
            (left) (256-512)$\times$(256-512)-pixel current map of monolayer MoS$_2$ visualized in RGB color-space. 
            (Center) corresponding binary pixel map showing every probe
    location;
            (right, inset) magnified unit cell.
    }
        % \textcolor{blue}{Need more detailed caption, and explain what is abcd subfigures are.}
    \label{fig:classic-flow}
\end{figure}
\begin{figure}[htbp]
    \centering
    \includegraphics[width=1.0 \textwidth]{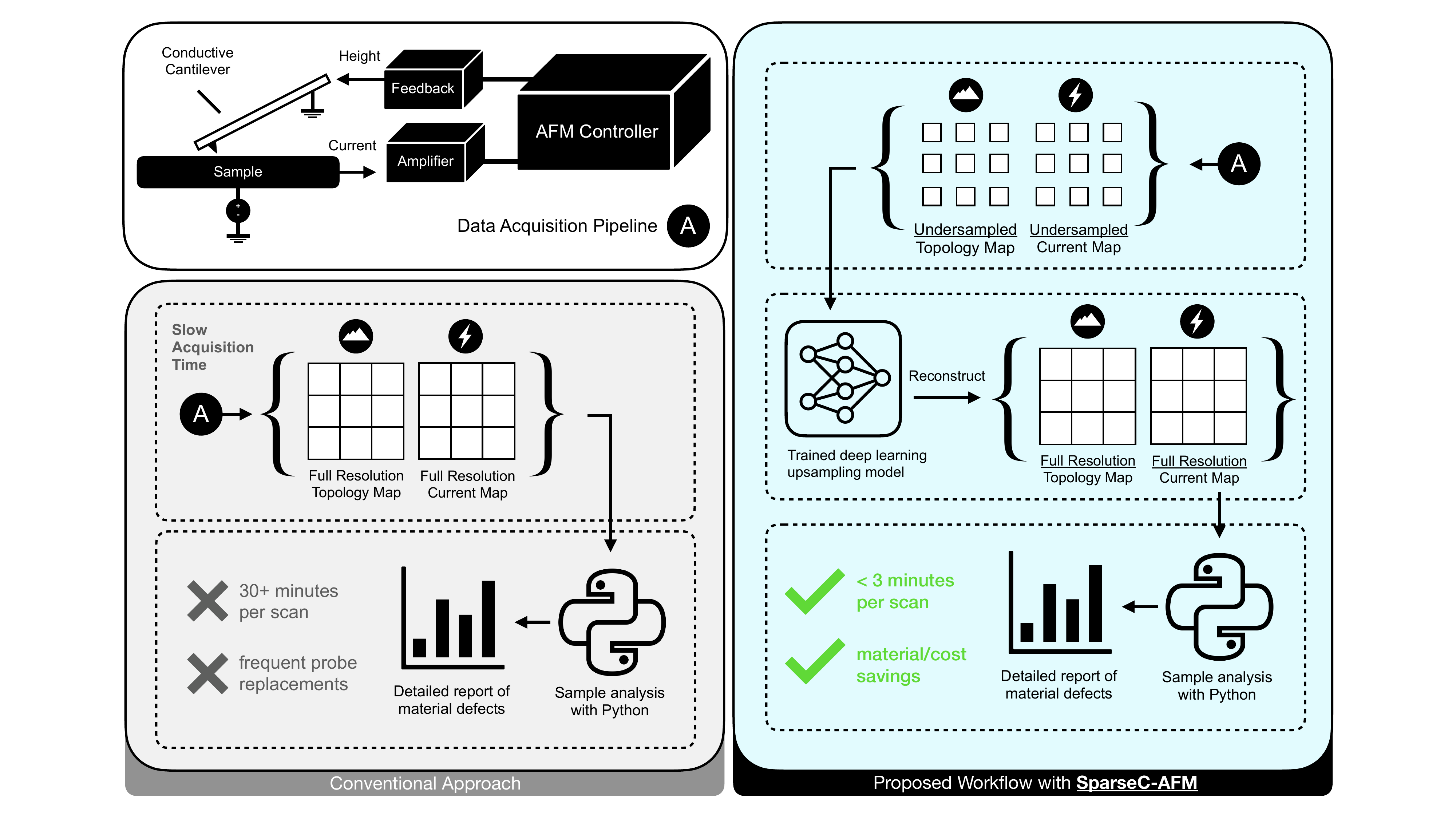}
    % \\[10pt]
    \caption{Our proposed workflow for rapid C-AFM scanning and characterization of 2D materials with \textbf{SparseC-AFM}. (Top-left) a high-level figure of our data acquisition pipeline for C-AFM. (Bottom-left) standard, full-resolution data collection and analysis procedure. (Right) our proposed, expedited workflow. A trained neural network predicts \underline{\textbf{full-resolution}} surface morphology and current maps from \underline{\textbf{undersampled inputs}}.
    }
    \label{fig:workflow}
\end{figure}

\subsection{An AI-Accelerated Workflow}

In recent years, artificial intelligence (AI) and machine learning (ML) have emerged as fields with transformative potential for materials science. These research areas have proposed end-to-end methods to extract complex features from high-dimensional datasets using large, statistical models. However, ``learning'' model parameters during the so-called ``training'' process typically requires large amounts of hand-labeled data. For example, training convolution neural networks (CNNs) to achieve high accuracy in tasks such as defect identification and material classification can require hundreds or thousands of labeled examples \cite{vasudevan2019materials, raghuraman2025imaging, liu2022experimental, yang2020automated, xing2018survey, wollmann2021deep, tek2009computer, linarto2023photon, dukes2023machine}. Curating training datasets incurs enormous time and labor costs. Thus, many AI-based techniques are impractical in highly dynamic environments such as semiconductor fabs. Moreover, AI models are usually trained with expensive hardware accelerators, further reducing their appeal for small labs \cite{laskar2023spm, chen2021acstem, rajagopal2023ai, kim2020machine, kraus2016hcs}.

In this paper, we present a deep learning model that rapidly and accurately reconstructs conductivity maps of 2D materials, such as MoS$_2$, from undersampled C-AFM scans. We compare (a.) classic flow implementation, where a high pixel-density C-AFM image (i.e., 15 min to collect) is parsed manually to extract relevant materials parameters, and (b.) our \textbf{SparseC-AFM} workflow, where identical operations are performed using an undersampled dataset (i.e., less than 5 min) and an AI model. Our novel workflow enables efficient extraction of critical 2D material parameters, including coverage percentage, the length and area of defect regions, and others. We achieved over \underline{\textbf{11$\times$ reduction}} in acquisition time compared to full-resolution C-AFM image analysis without significant degradation in data quality. Inspired by our promising results, we claim that this work represents a significant step for AI-assisted 2D material characterization, from laboratory research to industrial fabrication.
\section{Method}

\subsection{Data Collection}

We prepare samples of MoS$_2$ via plasma-enhanced chemical vapor deposition (PECVD) on Si-SiO$_2$ substrates. To train our upsampling model, we collect dual-channel (i.e., surface morphology and surface current) C-AFM scans of MoS$_2$ at a full resolution of 512$\times$512 (2 $\mu$m$\times$2 $\mu$m), and sparse resolutions of 256$\times$256, 128$\times$128, and 64$\times$64, using a commercially available Park NX-Hivac AFM system operated in contact mode.

We collect CAFM measurements under both high vacuum (~10$^{-5}$ torr) and ambient conditions with conductive PPP-CONTSCPt probes (nominal spring constant $k = 0.2$ N/m), manufactured by Nanosensors. An electrical back contact was established using eutectic silver (Ag) paste applied as a conductive adhesive on the substrate's backside, ensuring stable and consistent electrical measurements.

% Levi: we can resevere experiments with BTO for our upcoming letter; must likely out of scope for this paper
% Additionally, we capture topology maps of barium titanate (BTO) samples in contact mode. In total, we curate four datasets to train and evaluate our deep learning model.

\begin{figure}[htbp]
    \centering
    \includegraphics[width=1.0\textwidth]{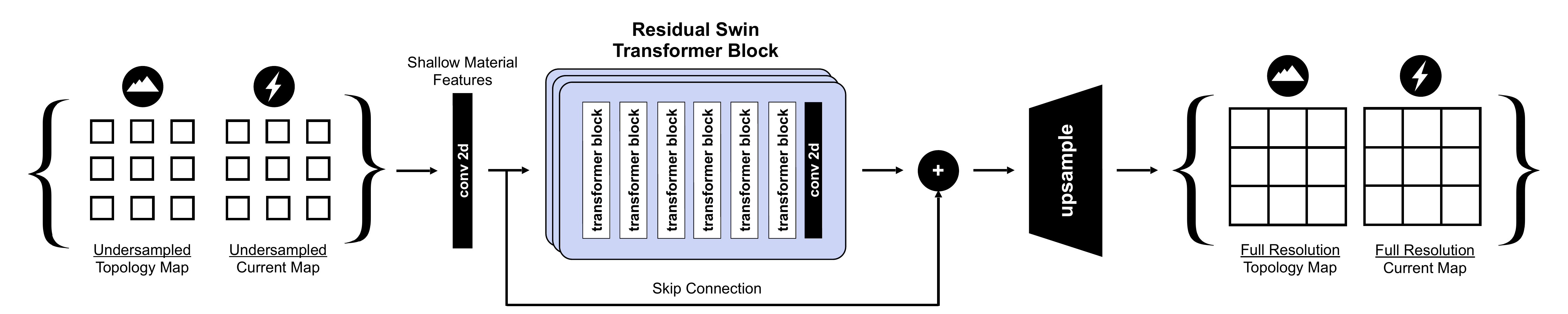}
    \\[7.5pt]
    \caption{A high-level diagram of our neural-network architecture. Our model, Sparse-C-AFM, can upsample surface morphology and current maps at multiple resolutions and levels of sparsity.}
    \label{fig:label-name}
\end{figure}

\subsection{Model Architecture}

To design our upsampling model, we take inspiration from the work of SwinIR~\cite{swinir}: a deep neural network designed for image restoration and super-resolution. Given a sparse, C-AFM scan of a material $X \in \mathbb{R}^{\frac{H}{\sigma} \times \frac{W}{\sigma}}$ where $X$ represents either a \textbf{surface-morphology} or a \textbf{surface-current} map, $H$ and $W$ are the original height and width, respectfully, and $\sigma \in \{2, 4, 8\}$ is the \textbf{upsampling factor}, we train a model $$f_{\theta}(\cdot) = \hat{y} \approx y,$$ where $\hat{y}, y \in \mathbb{R}^{H \times W}$, to map low-resolution inputs to high-resolution outputs.

Our model learns salient features of our samples using two stages: 1.) shallow feature extraction and 2.) deep feature extraction. Shallow feature extraction is handled by a single convolutional layer $H_{SF}(\cdot)$ $$F_{0}=H_{SF}(X).$$ 
Later, rich feature representations are extracted using stacked Swin Transformer blocks $H_{DF(\cdot)}$ $$F_{DF} = H_{DF}(F_0).$$ 
Our deep-feature extraction module leverages sifted-window attention initially proposed in Swin Transformer \citep{swin-transformer}, to enable a ViT \citep{ViT} backbone encoder to capture complex global dependencies without incurring high computational cost. Finally, we reconstruct full-resolution outputs via upsampling convolutional layers $H_{REC}$ $$\hat{y}=H_{REC}(F_0 + F_{DF}).$$ 
Our model is optimized via backpropagation using a standard pixel-wise loss $$\mathcal{L}=\| y - \hat{y} \|_{1}.$$
Readers are referred to our codebase for complete training and implementation details.

% By default, SwinIR is designed to process image-like inputs, (i.e., inputs with shape $\mathbb{R}^{C \times H \times W}$, where $C=3$). To leverage the pre-trained weights of this architecture, we pad our inputs to $\mathbb{R}^{C \times H \times W}$ by first adding an extra ''dummy" channel dimension, and repeating across the channel such that $C=3$.

\subsection{Use Case}

We integrate our upsampling model into a complete 2D material imaging and analysis pipeline designed for AFM (see Figure \ref{fig:workflow}). Initially, users collect a small amount of full-resolution data to finetune SparseC-AFM. This additional data allows our model to adapt to a novel distribution quickly and effectively. Once trained, users can scan samples at a sparsity of their choosing, using our fine-tuned model to produce full-resolution outputs from sparse inputs. We demonstrate the inherent tradeoffs between data sparsity and model accuracy in Figure \ref{fig:recon-hist}. We leave it to the reader to determine an acceptable threshold for these metrics.
\definecolor{custom_gray}{RGB}{241, 241, 241}
\sethlcolor{custom_gray}
\newcommand{\xmark}{--}

\section{Experiments \& Results}

\label{sec:Experiment}

\subsection{Model Training}

We train three variants of our upsampling model at $\times$2, $\times4$, and $\times$8 upsampling factors (i.e., $\times$4, $\times$16, $\times$64 total data reduction). Each model is trained on both surface morphology and current maps, using samples of MoS$_2$. We use the following architecture parameters: RSTB number 6, STL number 6, window size 8, channel dimension 180, and number of attention heads 6 \citep{swinir}. All models are trained for 200 epochs using 1024 steps per training epoch and batch sizes of 16, 16, and 4 for $\times$2, $\times4$, and $\times$8 upsampling factors, respectively. To improve the diversity of our training sets, we take random crops with side lengths of (256, 256, and 384) from (512$\times$512) full-resolution images for the $\times$2, $\times4$, and $\times$8 upsampling factors, respectively. We apply additional data augmentations to further boost model performance. All training runs are conducted on 1$\times$A6000 GPU using the Adam optimizer \cite{Adam} with a learning rate of 1e-5 and weight decay set to $\beta_1=0.9$, $\beta_2=0.99$.

% (see Figure \ref{tab:augmentations-psnr} for more details)

\subsection{Evaluation Framework}

\begin{figure}[bp]
    \centering
    \includegraphics[width=1.0 \textwidth]{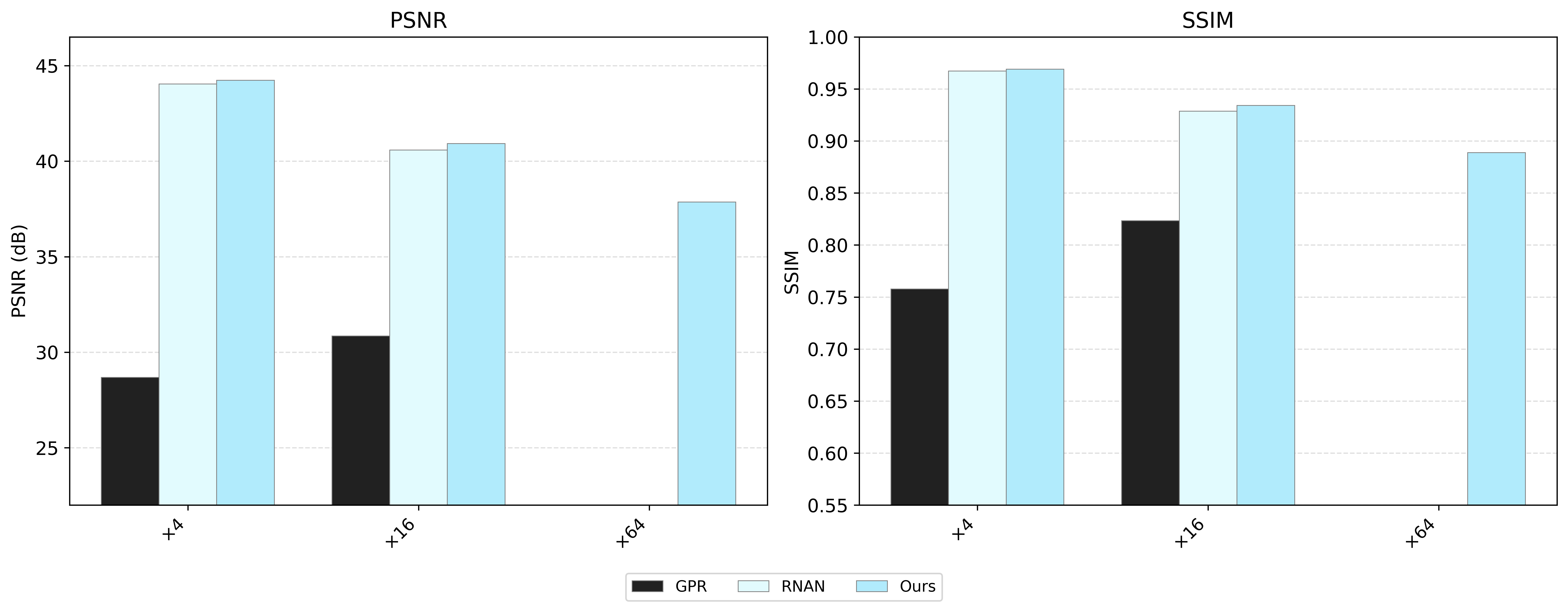}
    \caption{We compare our approach against previous sparse AFM reconstruction methods on common reconstruction metrics (i.e., SSIM \cite{Wang2004} and PSNR). *Gaussian Process Regression (GPR) has a time complexity of $\mathcal{O}(N^3)$ and is computationally infeasible for the ×64 sparsity case; therefore, these values are not reported. **RNAN does not support ×8 upsampling.}
    \label{fig:recon-hist}
\end{figure}

We evaluate the performance of our upsampling model across two axes. First, we apply traditional reconstruction metrics (e.g., PSNR, SSIM) to compare our approach directly against past works (see Figure \ref{fig:recon-hist}). Second, we employ a physics-informed Python script built using the \texttt{open-cv} library to extract electrical characteristics of original and model-predicted current maps. Our script maps full-resolution current maps to a set of scalar-valued material properties. Through extensive experimentation and evaluations, we demonstrate that SparseC-AFM not only outperforms previous AFM upsampling techniques but also that our model's outputs have \textbf{similar electrical properties} (i.e., film coverage, defect density, and localization of crystalline island boundaries, edges, and cracks) to full-resolution scans.

\subsection{Surface Morphology \& Current Map Upsampling}

\begin{figure}[t]
    \centering
    \resizebox{\linewidth}{!}{
        \setlength{\tabcolsep}{1pt}
        \begin{tabular}{llllll}
            \multicolumn{6}{l}{\textbf{Surface Morphology}} \\ [5pt]
            \includegraphics[width=0.20\linewidth]{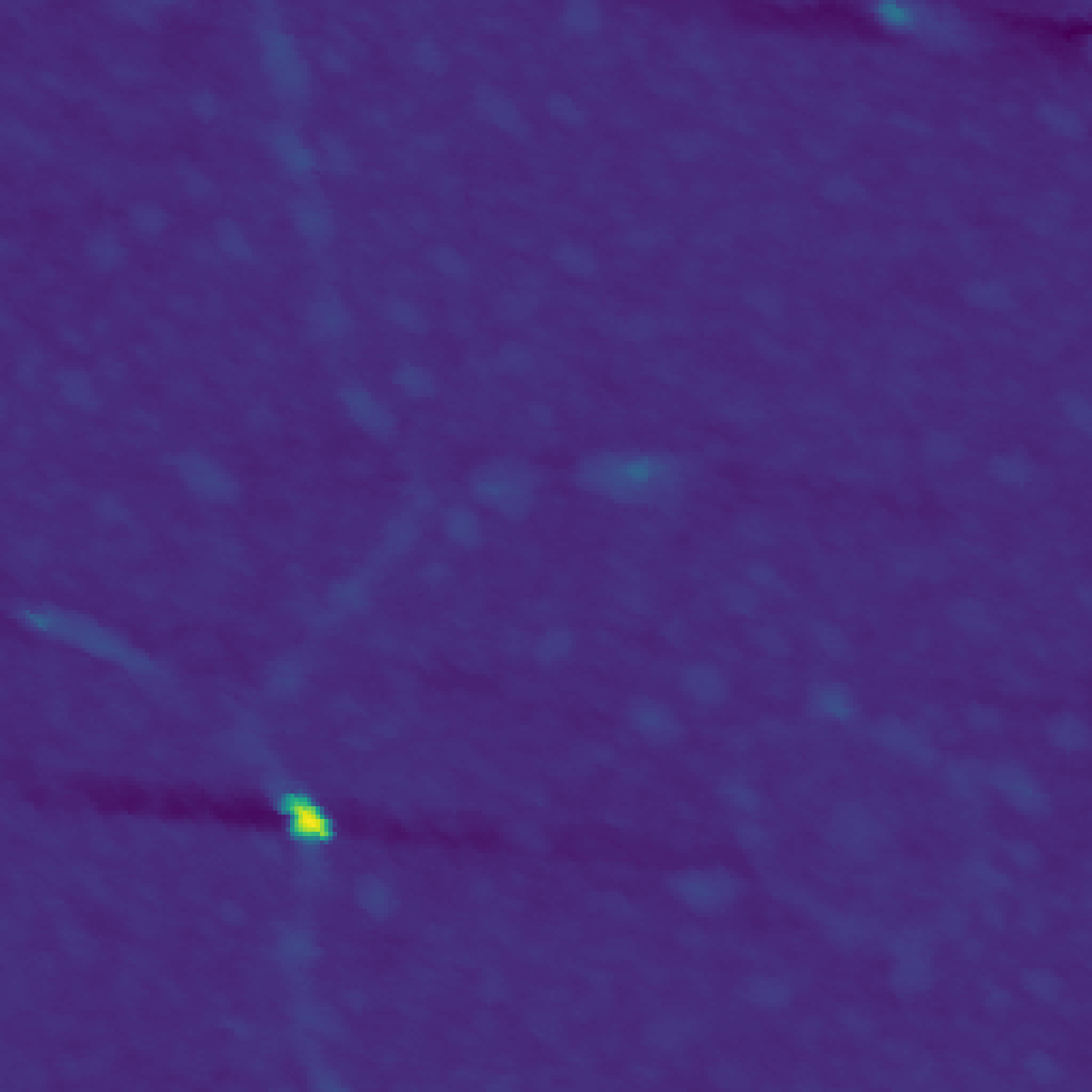} &
            \includegraphics[width=0.20\linewidth]{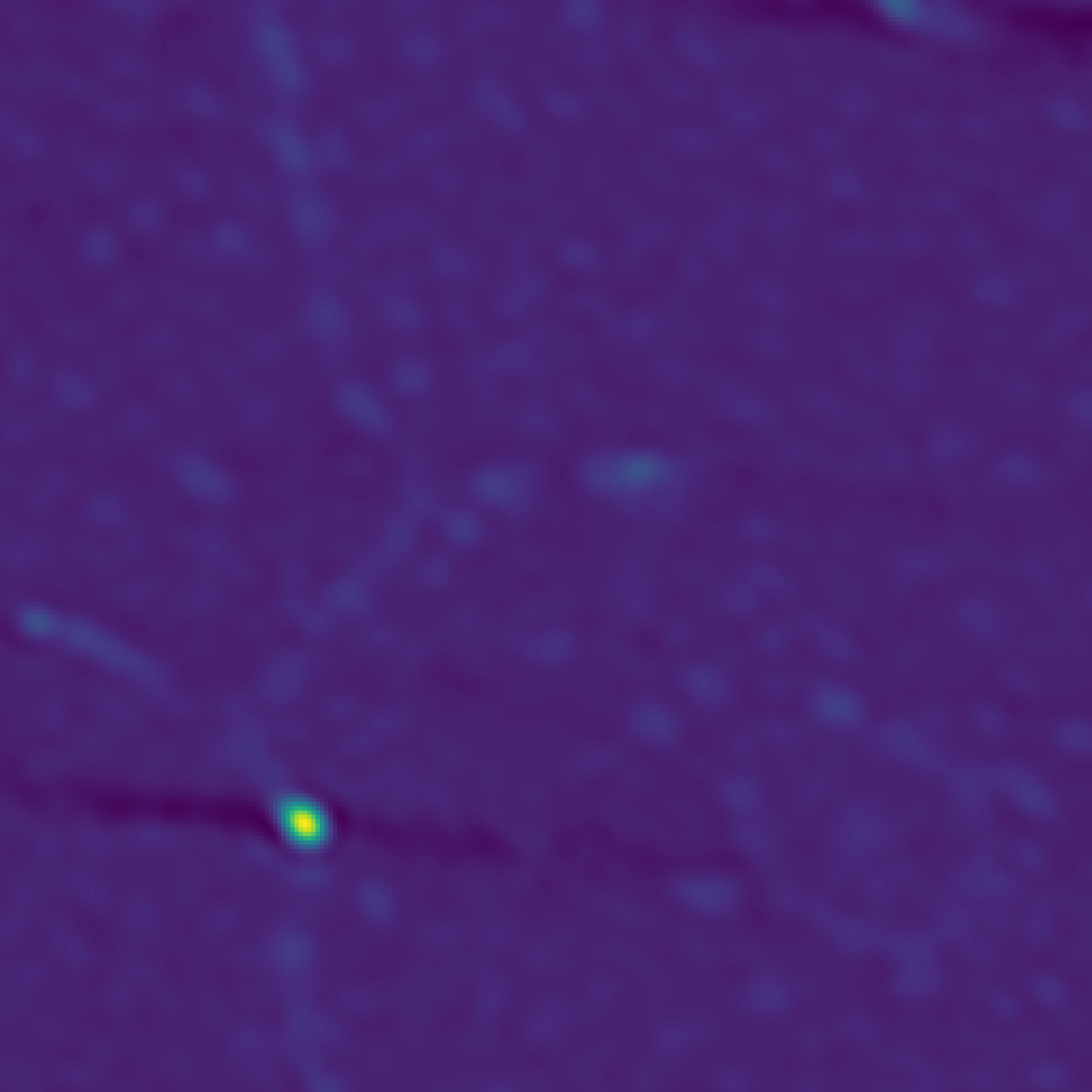} &
            \includegraphics[width=0.20\linewidth]{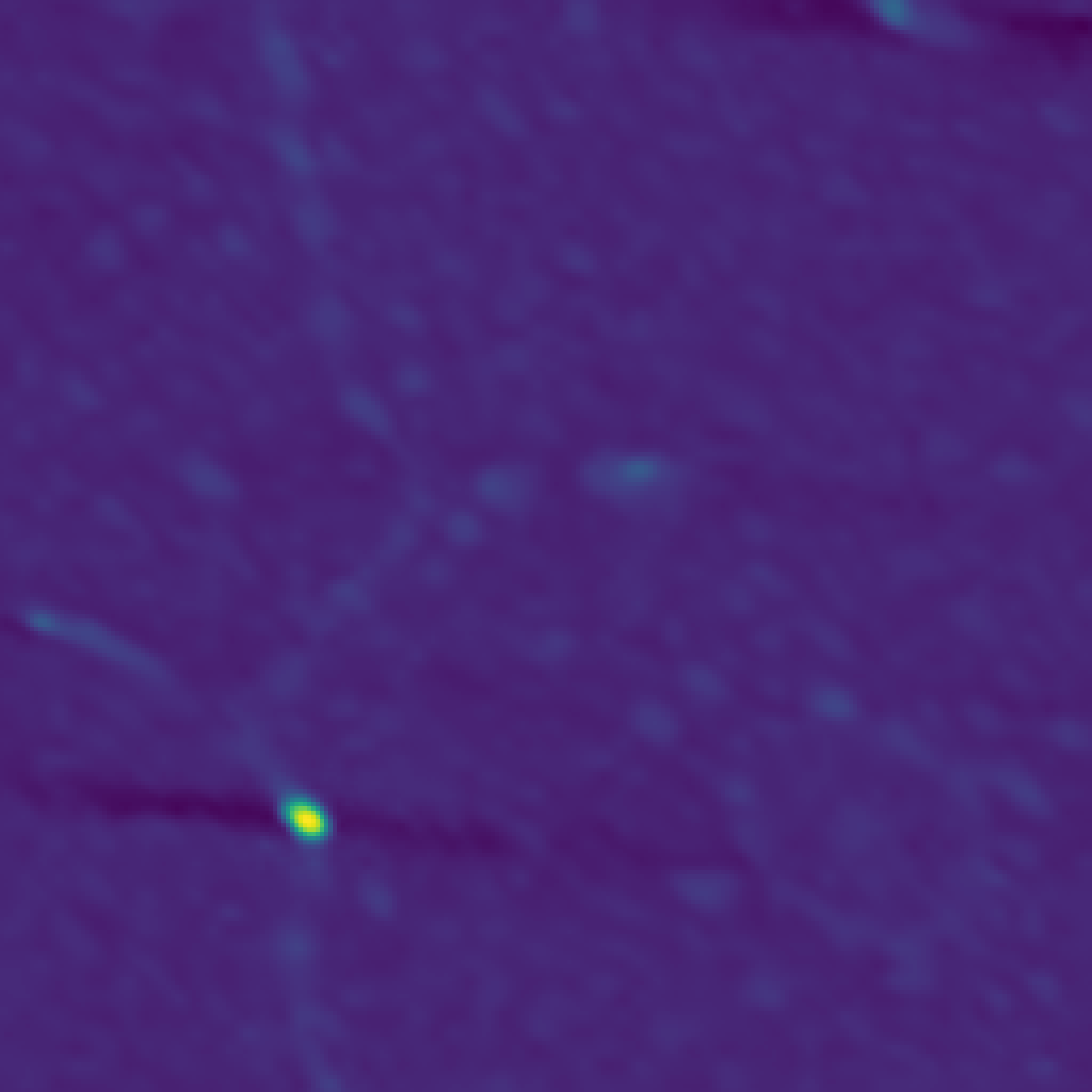} &
            \includegraphics[width=0.20\linewidth]{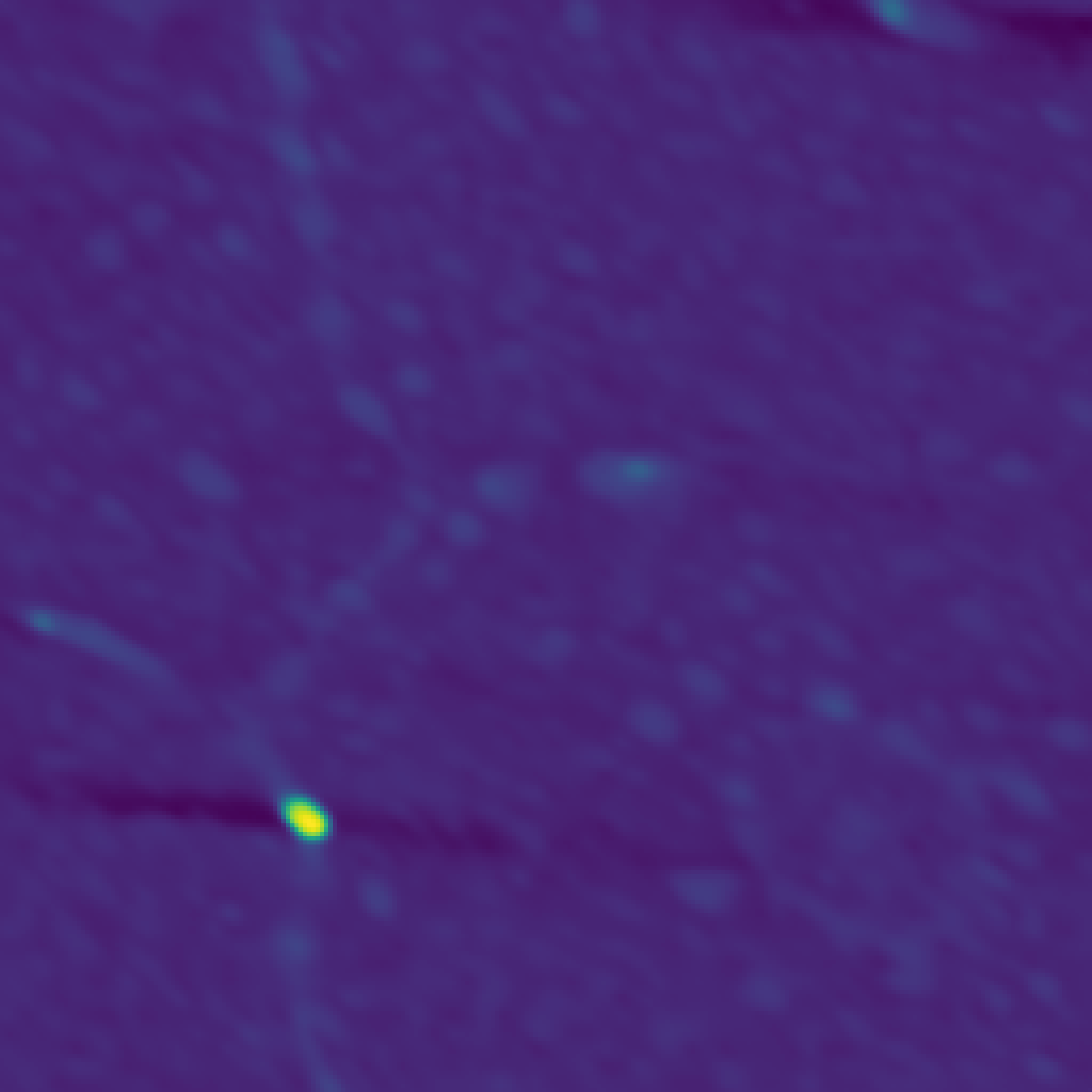} &
            \includegraphics[width=0.20\linewidth]{figs/qual-samples-two-row-grid/X/original.png} &
            \raisebox{-4.25pt}{\includegraphics[height=0.2145\linewidth]{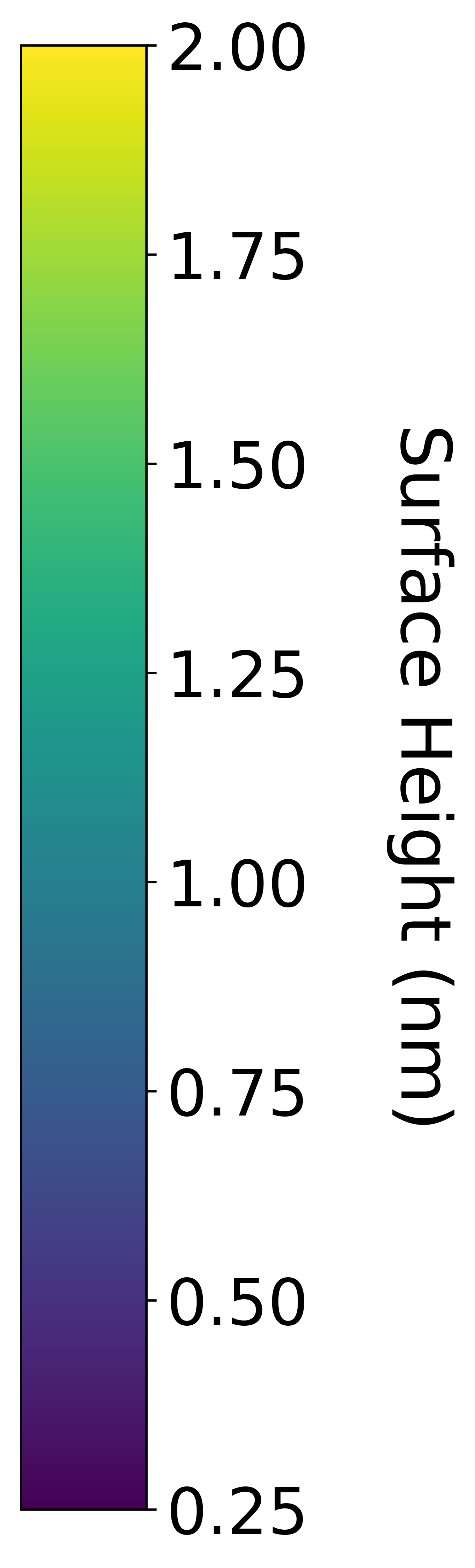}} \\ [5pt]
            \multicolumn{6}{l}{\textbf{Surface Current}} \\ [5pt]
            \begin{tabular}[t]{@{}c@{}}
                \includegraphics[width=0.20\linewidth]{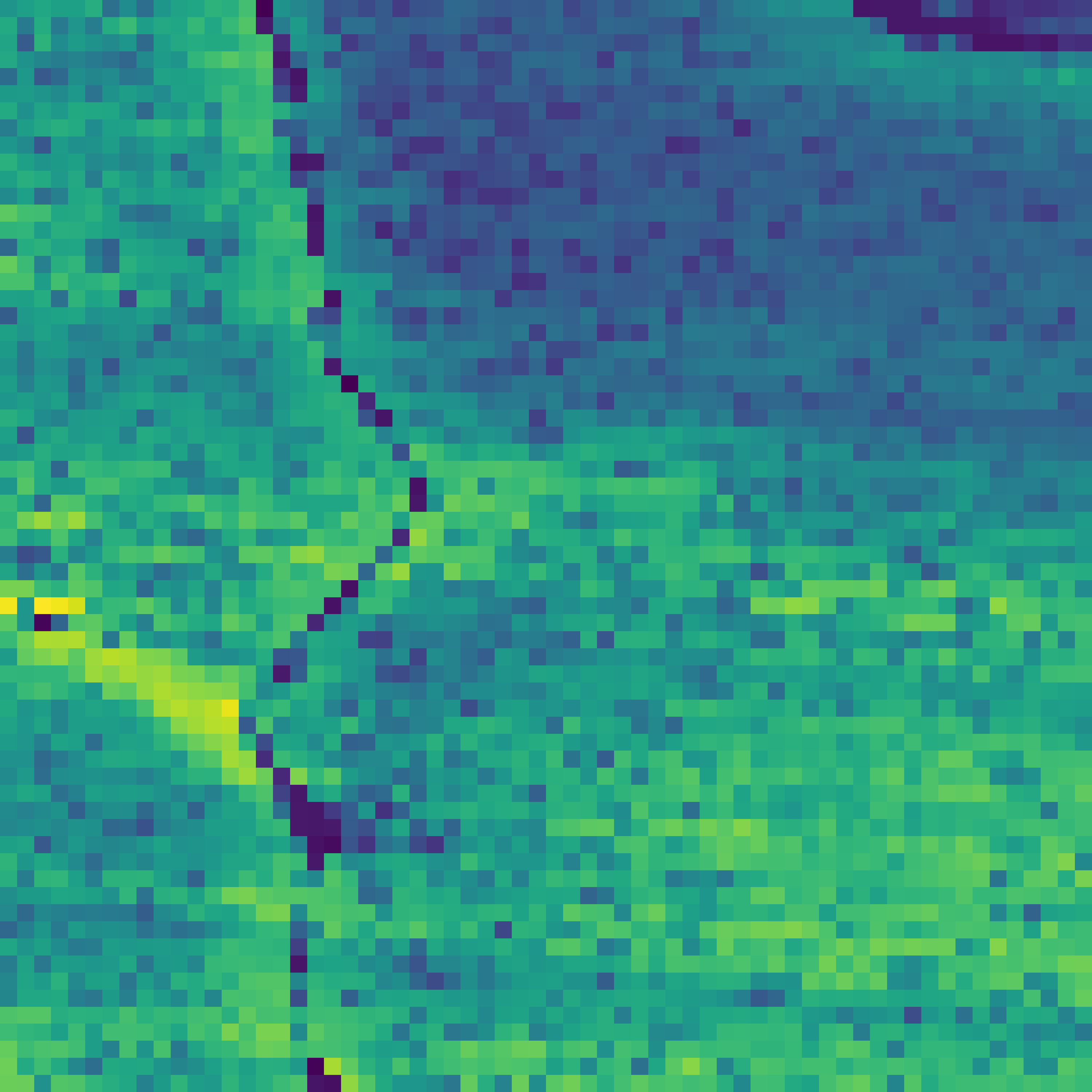} \\
                \small Sparse Input
            \end{tabular} &
            \begin{tabular}[t]{@{}c@{}}
                \includegraphics[width=0.20\linewidth]{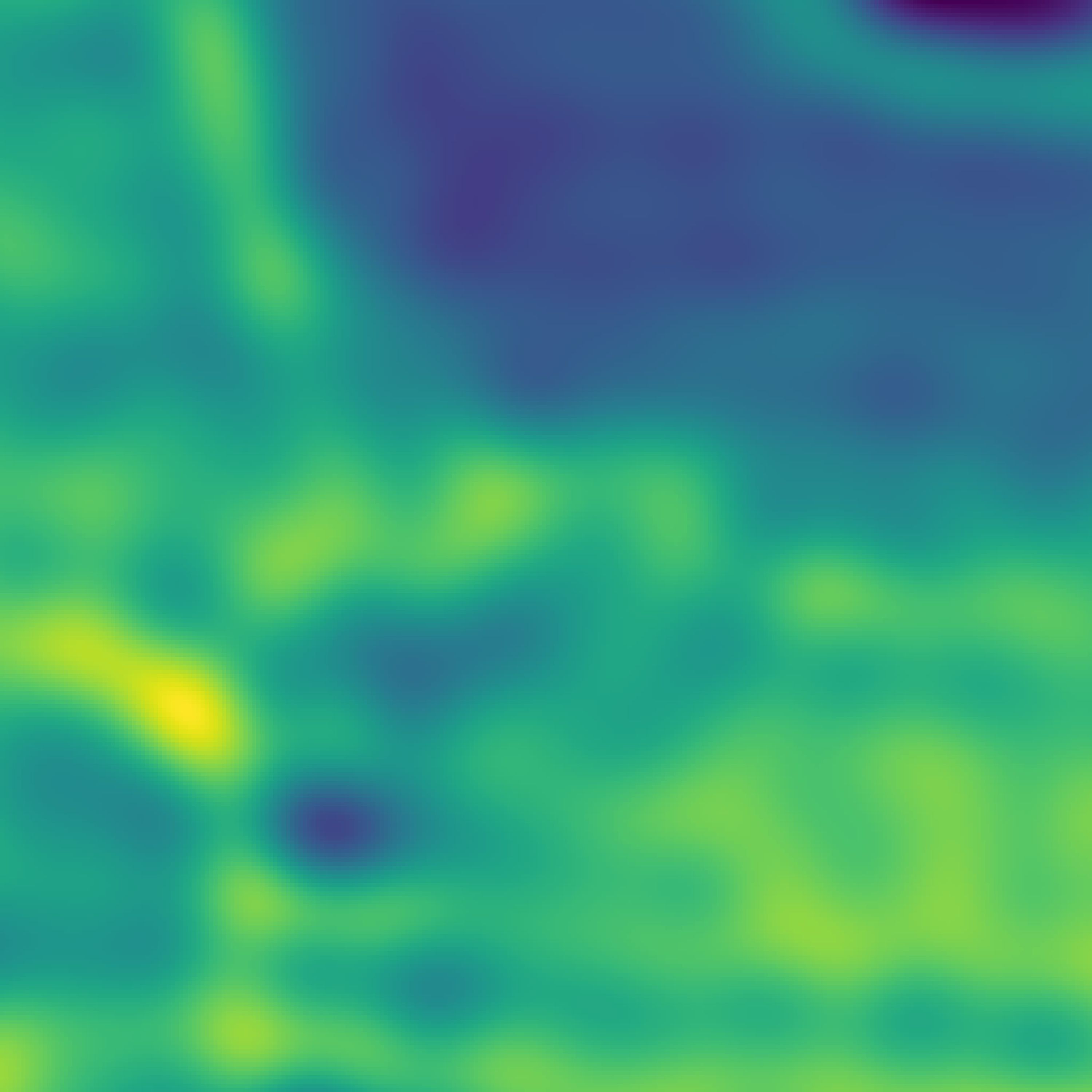} \\
                \small GPR \citep{Kelley2020-kt}
            \end{tabular} &
            \begin{tabular}[t]{@{}c@{}}
                \includegraphics[width=0.20\linewidth]{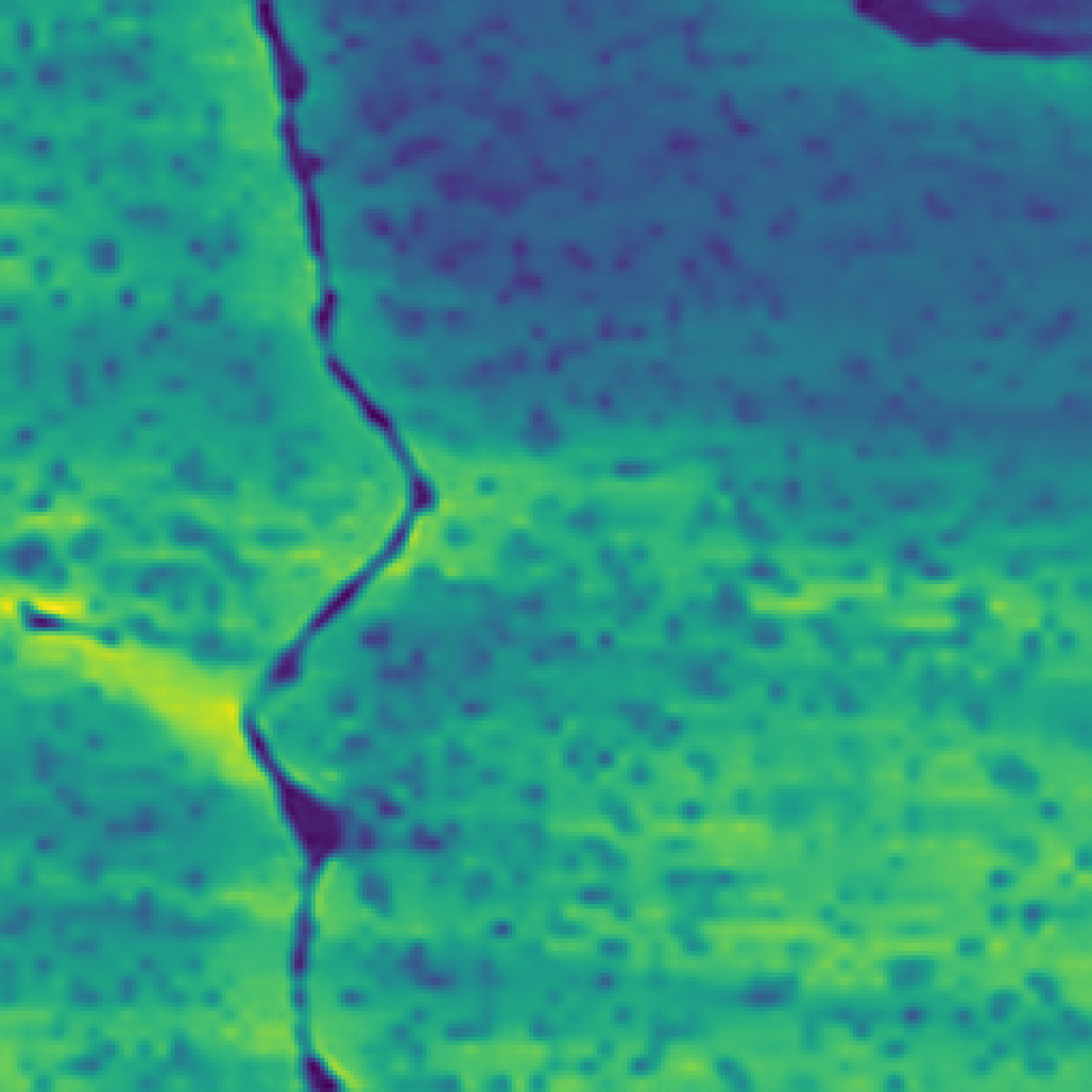} \\
                \small RNAN \citep{Kim2021}
            \end{tabular} &
            \begin{tabular}[t]{@{}c@{}}
                \includegraphics[width=0.20\linewidth]{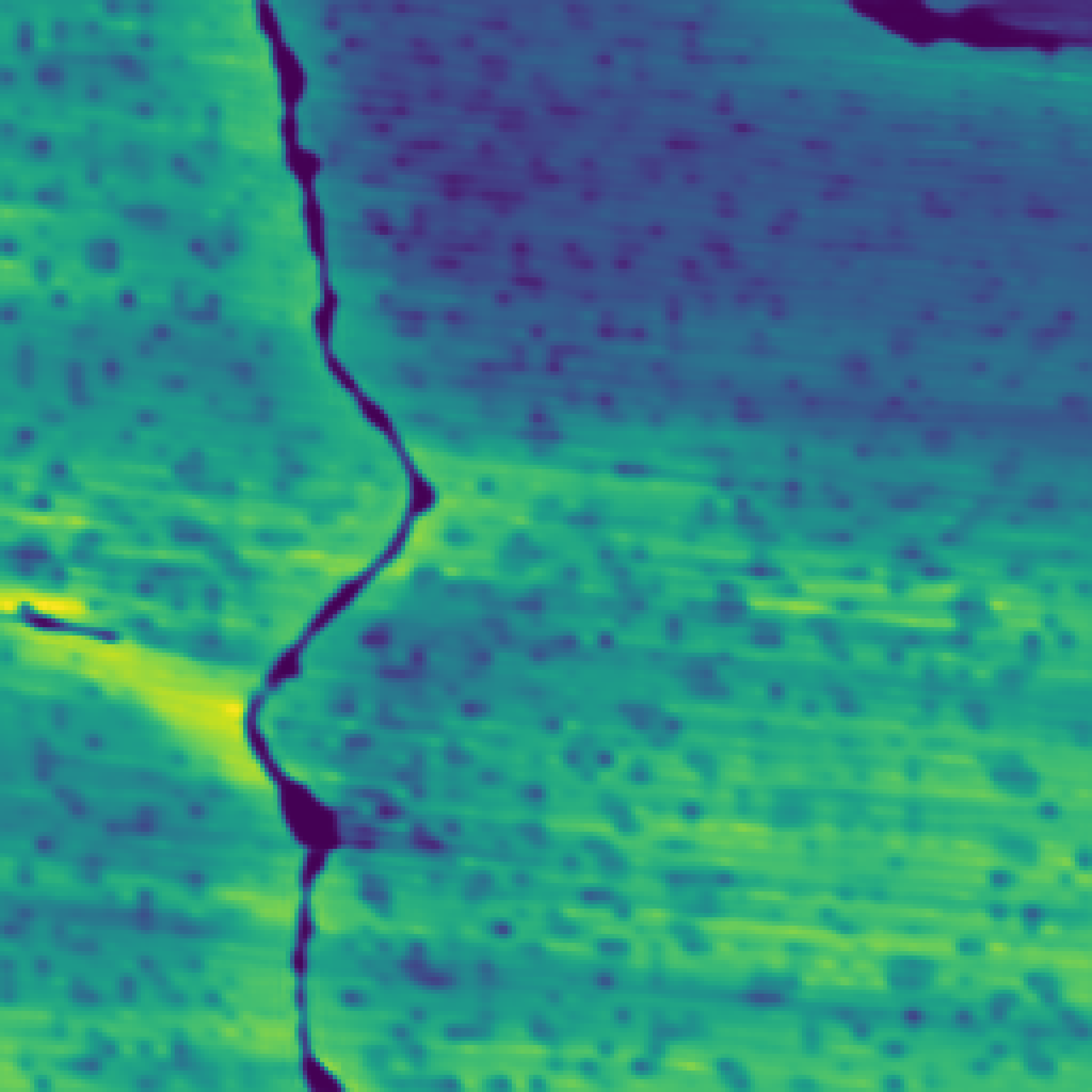} \\
                \small \textbf{SparseC-AFM} \\(ours)
            \end{tabular} &
            \begin{tabular}[t]{@{}c@{}}
                \includegraphics[width=0.20\linewidth]{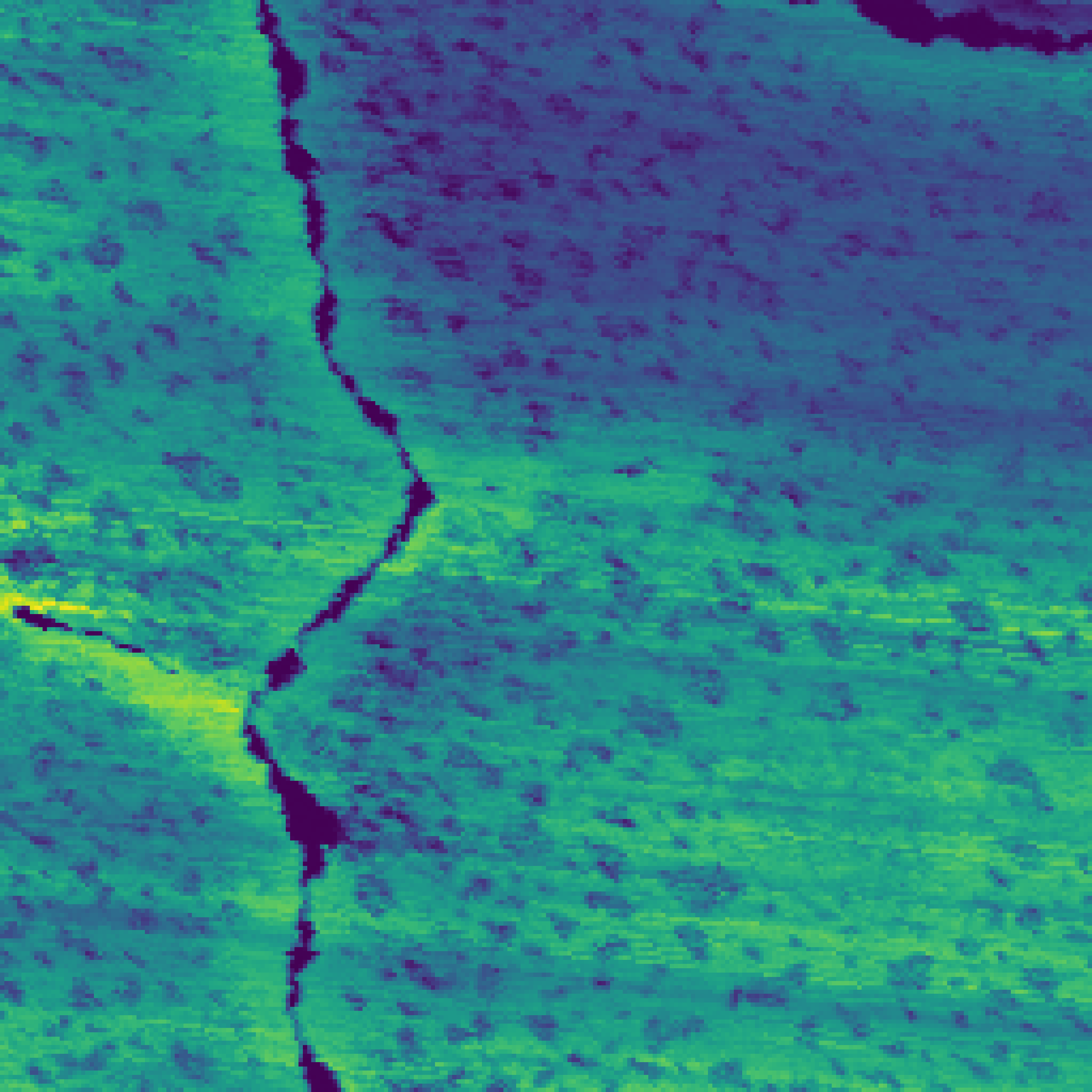} \\
                \small Original
            \end{tabular} &
            \raisebox{-4.25pt}{\includegraphics[height=0.2145\linewidth]{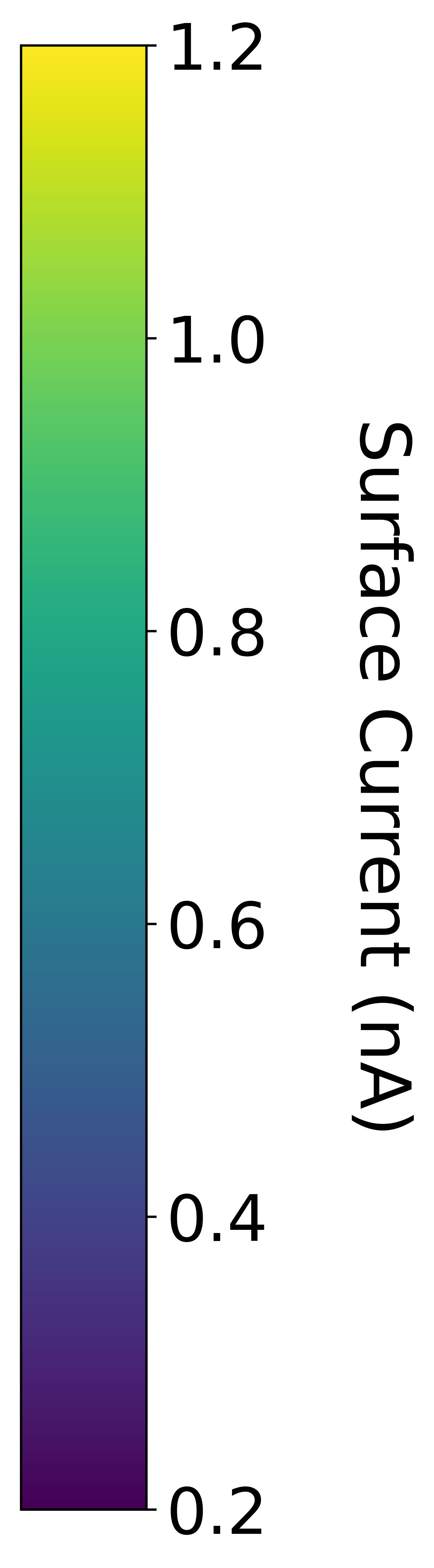}} \\ [5pt]
        \end{tabular}
    } % end resizebox
    \\[10pt]
    \caption{Qualitative results for $\times$4 upsampling of an \textbf{unseen} MoS$_{2}$ sample. The top and bottom rows represent surface morphology and current channels from the C-AFM mapping, respectively. All results are visualized in RGB color-space using the \texttt{matplotlib} Python package set to the \texttt{viridis} color mapping scheme.}
    \label{fig:grid-a}
\end{figure}

We share qualitative results for surface morphology and current map upsampling in Figure \ref{fig:grid-a}. Compared to previous approaches, SwinC-AFM produces outputs with higher characteristic fidelity and subjective ``sharpness". Moreover, our model generates predictions in a fraction of a second, whereas methods like GPR require retraining for novel inputs and take several minutes to run, even on high-end hardware.

\subsection{Upsampled Data Analysis \& Characterization}

% \begin{wrapfigure}{r}{0.30\textwidth}
%     \centering
%     \includegraphics[width=0.30\textwidth]{figs/transfer_learning.png}
%     \caption{We perform fine-tuning on a previously unseen dataset of barium titanate (BTO) at $\times$2 super-resolution. Our model demonstrates excellent few-shot capabilities with minimal data.}
%     \label{fig:transfer-learning}
% \end{wrapfigure}

% \input{fig_tex/electric_properties_t1}
\begin{figure}[htbp]
    \centering
    \includegraphics[width=0.80 \textwidth]{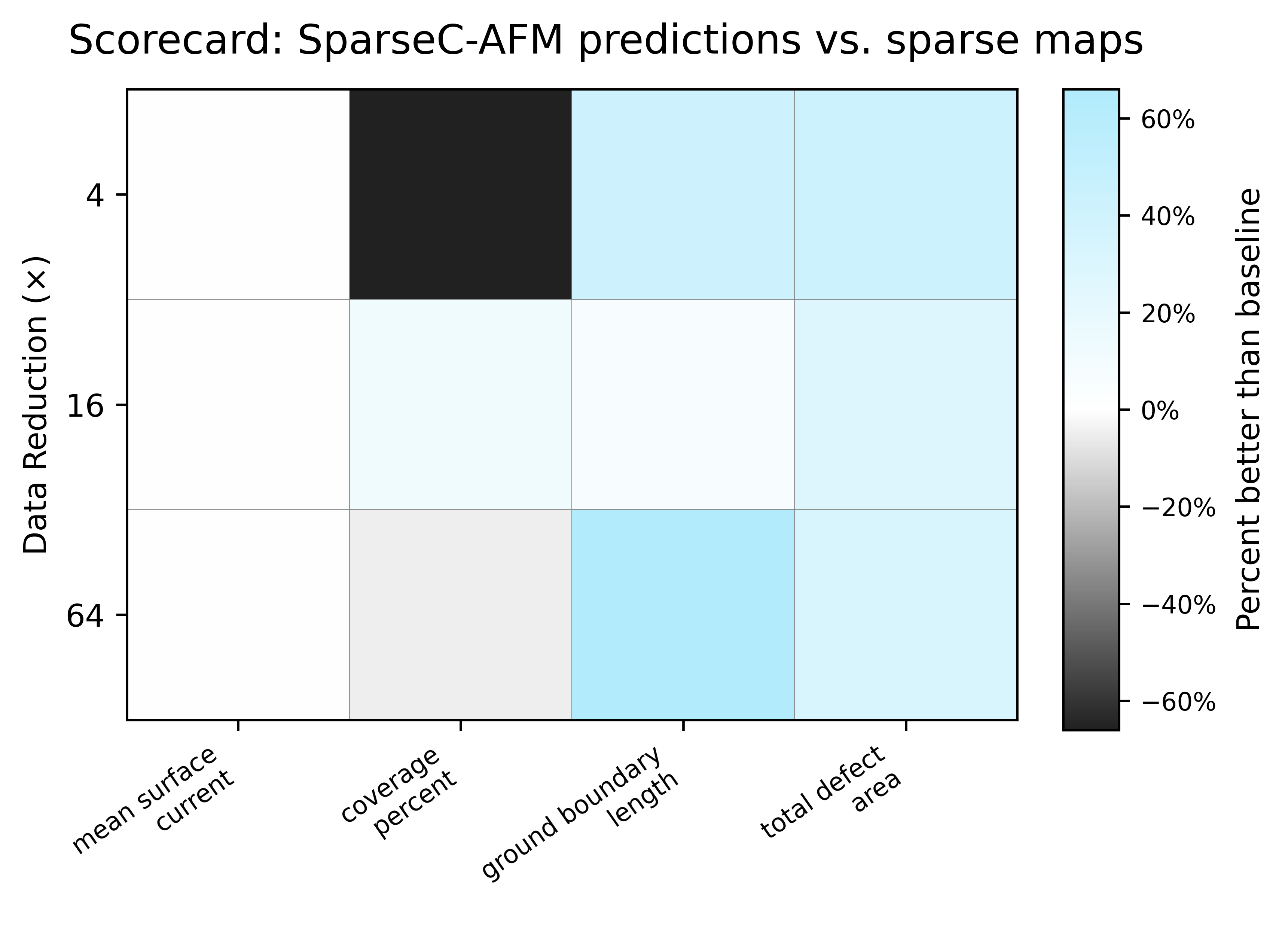}
    \caption{Are SparseCAFM-upsampled current maps true to life? We compare the electrical characteristics (e.g., mean surface current, coverage percentage, etc) of sparsely sampled MoS$_2$ to SparseCAFM predictions at varying levels of sparsity ($\times 4$, $\times 16$, $\times 64$). The scorecard figure above reports the relative, mean absolute errors of SparseCAFM predictions and sparse current maps (baseline) using ground-truth data collected using classic-flow C-AFM scanning. For example, area extended shape values of SparseCAFM-upsampled current maps are over 60\% more accurate than our baseline using a total data reduction factor of $\times 64$.}
    % \textcolor{blue}{add more detailed caption to explain what it is and what is the conclusion from this figure.}
    \label{fig:classic-flow}
\end{figure}

Detailed sample analysis and characterization are critical stages in many industrial processes. To develop a holistic picture of our model's performance, we evaluate outputs using traditional reconstruction metrics and electrical properties of interest. We utilize a Python script to extract salient material properties from full-resolution inputs. Using this tool, we confirm that our model's predicted current maps have \textbf{similar electrical characteristics} to ground truth data. We report results for an extensive suite of electrical properties in Figure~\ref{fig:classic-flow}.

% \textcolor{blue}{Please check whether this figure link is correct.}

% \input{fig_tex/augmentations_t1}

% Levi: I believe these claimns are out of scope and weakly supported; removed
% \subsection{Deployment on Novel Datasets}
% Users can quickly re-train SwinC-AFM and deploy on novel datasets, experimental conditions, and materials. To demonstrate the generality of our approach, we conduct a series of experiments measuring our model's performance on unseen datasets. Here, we train three models at $\times$2, $\times$4, and $\times$8, upsampling factors using identical training parameters used in previous runs. Figure \ref{fig:transfer-learning} elucidates the real-world of our work, showing our model's strong performance in few-shot, even \textbf{zero-shot} scenarios.

% Levi: These PSNR values seem unbelieveably high to me...
% \subsection{Ablations}
% In Table \ref{tab:augmentations-psnr}, we validate the inclusion of different data augmentation techniques. Here, we train separate variants of our model with an upsampling factor of $\times$2 and all other training parameters set identically as before.
\section{Discussion}
\label{sec:Conclusion}

We present \textbf{SparseC-AFM}, a deep learning-based workflow for AFM that accelerates imaging and study of 2D materials. Unlike previous fast AFM techniques, our proposed method is both \textbf{1.) non-intrusive} and \textbf{2.) highly generalizable}, allowing rapid development, failure analysis, and characterization of various 2D materials. Our method is 1.) non-intrusive as it does not require major changes in data collection or analysis procedures. Furthermore, our approach is 2.) highly generalizable because it adapts quickly and effectively to unseen datasets and materials, with varying levels of sparsity across different experimental conditions. Inspired by large reductions in data acquisition time and cost, we claim that our work represents a significant step forward from laboratory analysis to industrial production of 2D materials.

% \subsection*{Acknowledgments}
% ...


\bibliographystyle{spiebib} % makes bibtex use spiebib.bst
\bibliography{main} % bibliography data in report.bib

\end{document}